\pdfoutput=1

\documentclass[11pt]{article}

\usepackage{acl}

\usepackage{times}
\usepackage{latexsym}

\usepackage[T1]{fontenc}

\usepackage[utf8]{inputenc}

\usepackage{microtype}

\usepackage{amsmath}
\usepackage{amssymb}
\usepackage{pifont}
\usepackage{adjustbox}
\usepackage{hyperref}
\usepackage{subcaption}
\usepackage{caption} 
\usepackage{fixltx2e}
\usepackage{multirow}
\usepackage{stfloats}
\usepackage{xurl}

\usepackage{booktabs}

\usepackage{algorithm,algpseudocode}

\title{RoViST: Learning Robust Metrics for Visual Storytelling}

\author{
Eileen Wang\qquad
Soyeon Caren Han\thanks{\, Corresponding author (caren.han@sydney.edu.au)}\qquad
Josiah Poon
\\
School of Computer Science, The University of Sydney, NSW, Australia \\
\texttt{\{Eileen.Wang, Caren.Han, Josiah.Poon\}@sydney.edu.au} \\
  }

\begin{document}

\setlength{\abovedisplayskip}{5pt}
\setlength{\belowdisplayskip}{5pt}

\maketitle
\begin{abstract}
Visual storytelling (VST) is the task of generating a story paragraph that describes a given image sequence. Most existing storytelling approaches have evaluated their models using traditional natural language generation metrics like BLEU or CIDEr. However, such metrics based on $n$-gram matching tend to have poor correlation with human evaluation scores and do not explicitly consider other criteria necessary for storytelling such as sentence structure or topic coherence. Moreover, a single score is not enough to assess a story as it does not inform us about what specific errors were made by the model. In this paper, we propose 3 evaluation metrics sets that analyses which aspects we would look for in a good story: 1) visual grounding, 2) coherence, and 3) non-redundancy. We measure the reliability of our metric sets by analysing its correlation with human judgement scores on a sample of machine stories obtained from 4 state-of-the-arts models trained on the Visual Storytelling Dataset (VIST). Our metric sets outperforms other metrics on human correlation, and could be served as a learning based evaluation metric set that is complementary to existing rule-based metrics.\footnote{The RoViST code: \url{https://github.com/usydnlp/rovist}}
\end{abstract}

\section{Introduction}
\begin{figure}[t!]
  \centering
  \includegraphics[width=1\linewidth]{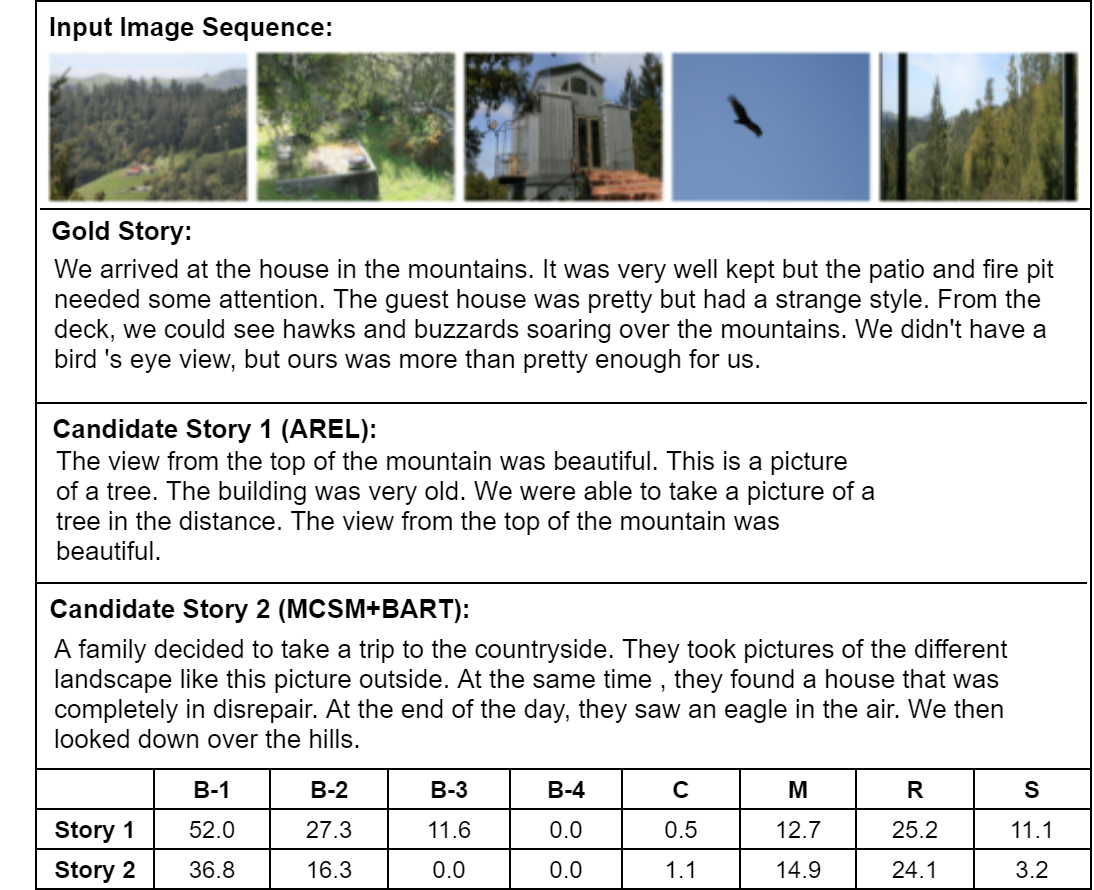}
  \caption{Example gold story found in the VIST dataset versus machine output from 2 VST models and their $n$-gram based metrics.}
  \label{fig: Story 2}
\end{figure}

Visual storytelling (VST) is a natural language generation (NLG) task that aims to automatically generate a cohesive story given a sequence of images \citep{huang2016visual}. The task is fundamental to the development of intelligent agents capable of understanding complex visual scenarios, and can be further applied to assist the visually impaired in understanding images on the web. Recently, progress has been made on designing network architectures to accomplish the VST task but little work has been done to explore new metrics that automatically evaluate and quantify the errors produced by these systems. As to date, a majority of the past works on VST have used existing popular \textit{n}-gram based metrics such as BLEU, METEOR, ROUGE, CIDEr, and SPICE to evaluate their models \citep{wang2018no, kim2018glac, hsu2019visual, chen2021commonsense}. However, it is known that such metrics are unreliable for VST. Figure \ref{fig: Story 2} shows two machine generated stories for a photo sequence and their corresponding \textit{n}-gram matching based metrics (BLEU, CIDEr, METEOR, ROUGE-L and SPICE). Evidently, the first candidate story is more repetitive and lacks a narrative style but achieves higher scores across a majority of the $n$-gram based metrics in Figure \ref{fig: Story 2}. The second story however, has greater word diversity and is more expressive through its use of phrases like `\textit{completely in disrepair}'. Relevant words like `\textit{trip}', `\textit{countryside}' and `\textit{hills}' are also used but are not rewarded since they are not mentioned in the gold story.

The low level of agreement between human judgement and current automatic metrics may be because such metrics were originally developed to assess machine translation, summarization and image captioning tasks \citep{sharif2018nneval}, which are significantly different problems to VST. Specifically, VST is a multimodal task that firstly requires: 1) generating text relevant to the image content but unlike image captioning, there is less emphasis on describing relationships between objects and may contain concepts that are inferred from the image. It additionally needs to ensure that: 2) the story must be topically coherent, similar to how a human would tell a story in a social setting. Sentences should not sound disjointed e.g. `\textit{We went to the park. I grew up in Sydney}'. And finally 3) avoids repetition which appears to be a common issue in current VST models. For instance, Candidate Story 1 in Figure \ref{fig: Story 2} exhibits inter-sentence repetition between the first sentence and last sentence. We also find that some output stories may contain repetition \textit{within} sentences (i.e. intra-sentence repetition) e.g. `\textit{we had a good time and had a great time!}'. 

Moreover, it is noted that open-ended text generation tasks usually suffer from the one-to-many issue, whereby there are multiple plausible outputs for the same input which are not fully reflected in the reference sentences \citep{guan2020union}. This issue is even more prominent in the VST task as different individuals may tell significantly different stories and have diverse interpretations given the same image sequence. All these issues suggest that we require evaluation metrics that do not simply rely on comparison with reference sentences. In addition, given that the VST task requires several aspects, one single metric is not sufficient to evaluate a story and there is a need to design multiple interpretable metrics that each target a specific VST criteria. Hence, in this paper, we propose several unreferenced metrics for the VST task based on the three aforementioned criteria: 1) visual grounding, 2) coherence, and 3) non-redundancy. 

 To address criteria 1), we propose a learned metric to calculate relevance scores between nouns in the VST sentences with the bounding box regions in the images. We decide to focus on nouns as they provide the most visual information. Other words like adjectives and adverbs are difficult to ground and such words may differ significantly depending on the person writing the story. The second criteria which is story coherence requires that consecutive sentences flow and that each sentence is not just an isolated description of the image. Existing methods for measuring coherence have used next sentence prediction (NSP) to find the probability that a sentence comes after a preceding sentence \citep{hu2020makes}. Inspired by this method, we fine-tune the ALBERT \citep{lan2019albert} model on story sentences and build a sentence-order prediction (SOP) model. Finally, to address criteria 3), we propose an additional metric to explicitly measure inter-sentence and intra-sentence repetition.

The contributions are summarized as follows: 
1) We propose an interpretable and reference-free metric that addresses 3 criteria required for VST - visual grounding, coherence and non-redundancy. 2) We conduct human evaluation studies to assess a sample of machine generated stories obtained from 4 state-of-the arts VST models. 3) We test the effectiveness of our proposed metrics by analyzing its correlation with human scores and show that our metrics outperform other existing metrics that are commonly used for VST and NLG tasks.

\section{Related Works}
\textbf{Natural Language Generation Metrics} 
The most popular NLG evaluation metrics are BLEU \citep{papineni2002bleu}, ROUGE \citep{lin2004rouge}, METEOR \citep{banerjee2005meteor}, CIDEr \citep{vedantam2015cider} and SPICE \citep{anderson2016spice}. All these metrics are widely used in evaluating image captioning tasks \citep{anderson2018bottom, zhou2020unified} and have also been predominantly used in VST tasks \citep{wang2018no, hsu2019visual, chen2021commonsense} due to the lack of metrics designed for VST. While these metrics are computationally efficient, they have limited ability in accounting for synonym matches or phrase reordering. This poses a problem for many open-ended text generation tasks like VST where different annotators may have slightly different (but still plausible) ways of describing the same image. To address this, some metrics focus on comparing distance and similarity between word embeddings such as Word Mover's Distance \citep{kusner2015word}, MoverScore \citep{zhao2019moverscore} and  BERTScore \citep{zhang2019bertscore}. However, these metrics mentioned so far still heavily rely on similarity with references, potentially leading to bias for VST tasks as the references may not fully cover the possible ways to write a story for an image sequence.

\textbf{Visual Grounding Metrics}
Past studies have proposed examining the images in addition to human written references. \citet{cui2018learning} trained a binary classifier to discriminate between human and machine captions using image and text representations obtained from a CNN and RNN. TIGEr \citep{jiang2019tiger} employs the pretrained SCAN model \citep{lee2018stacked} to calculate the text-to-image grounding scores and compares the relevance ranking and grounding weights distribution among image regions between the references and the candidate. \citet{lee2020vilbertscore} later introduced ViLBERTScore which uses the same approach as BERTScore but utilizes the ViLBERT model \citep{lu2019vilbert} to retrieve image-conditioned token embeddings. However, we note that these methods are initially designed for evaluating image captioning systems. Hence, while they do consider the text-to-image similarity aspect, they do not explicitly address the extra criteria required for VST such as story coherence. Moreover, such metrics still rely on reference sentences to some extent. 


\textbf{Story Generation Metrics}
Language models like BERT \citep{devlin2018bert} trained with NSP and masked language modelling tasks can identify appropriate use of words and sentences and hence, may show promising results when applied to evaluating open-ended text generation. \citet{guan2020union} proposed UNION, an unreferenced metric for scoring machine generated stories. They leverage a BERT model trained with negative samples created by perturbing ground truth stories and predicts a score representing how human-like a story is. They showed the effectiveness of BERT in identifying stories with conflicting logic, repeated plots and incoherence. However, UNION purely evaluates the output text and cannot be applied to analyse the text-to-image relatedness required for the VST multimodal task. Additionally, a single score is outputted which is not informative enough to gauge what specific errors were made by the model.
Moving to VST, \citet{hu2020makes} designed reward functions to capture story quality for VST models that use a reinforcement learning framework based on 3 criteria: image relevance, coherence and expressiveness. Image relevance is measured by $n$-gram precision of entities between candidate and reference sentences, coherence through BERT's NSP task, and word diversity by computing BLEU scores between generated sentences. 

Inspired by this, we also analyze story quality from 3 similar perspectives 1) visual grounding, 2) coherence, and 3) non-redundancy. We attempt to extend the methods of \citet{hsu2019visual}, provide a reference-free approach and conduct a comprehensive analysis with human evaluation.

\section{Method}
We describe our proposed metric in detail. Given a machine story, we aim to output 3 scores that explicitly evaluates the story based on 1) visual grounding, 2) coherence, and 3) non-redundancy. 

\subsection{RoViST-VG: Visual Grounding Scorer}
\begin{figure}[t!]
  \centering
  \includegraphics[width=1\linewidth]{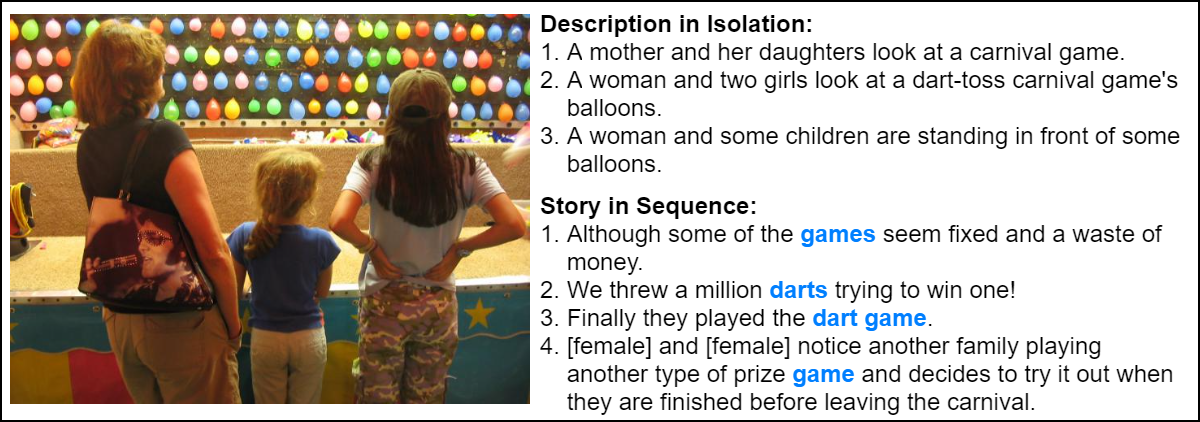}
  \caption{Example ground truth description in isolation (dii) and story in sequence sentences (sis) sentences corresponding to an image from the VIST dataset.}
  \label{fig: dii vs sis}
\end{figure}

To detect the visual relationship between image and text, we build a model that computes the similarity between the nouns in the story sentences with the bounding box regions in the images. We focus specifically on nouns because despite the diverse range of words one can use when storytelling, we notice that the main commonality among the ground truth sentences is the noun mention. This is most likely because nouns (in particular, tangible nouns) tend to offer the most visual information and is the common element that people would recognize when observing an image. An example of this case is in Figure \ref{fig: dii vs sis} where we can see that the nouns `\textit{dart}' and `\textit{game}' tends to appear in multiple gold sentences, even though each sentence is quite different in structure. 

Our visual grounding scorer is inspired by the phrase localization task \citep{plummer2015flickr30k} which involves learning to align sentence entities with image regions. We note that we could have just employed typical image-text matching models like SCAN \citep{lee2018stacked} to calculate a similarity score between image and text. However, such models are trained on image captioning sentences and do not explicitly focus on the more fine-grained task of word-region alignment. Moreover, retraining these models with VIST images and whole sentence pairs would be challenging as previously mentioned, story sentences tend to differ significantly in semantics and structure due to human imagination. This is in contrast to image captions where ground truth sentences typically tend to be similar to each other even across different human annotators (e.g. see description in isolation sentences in Figure \ref{fig: dii vs sis}). 

Inspired by CLIP \citep{radford2021learning}, we create a model that learns the image region and text embeddings such that the noun mention corresponding to an image region will have similar vector representations in geometric space. Let $I_i$ be an image of a bounding box region and $T_i$ be the matching noun. For the image encoder, we follow \citet{radford2021learning} and leverage the Vision Transformer (ViT) \citep{dosovitskiy2020image} to first extract the image features from $I_i$. An additional linear head is further added to project the features to a vector embedding of dimension 1024. For the text encoder, $T_i$ is first converted to 300 dimensional GLoVe vectors \citep{pennington2014glove}. If $T_i$ is composed of more than one word, the GLoVe vectors of each token are simply averaged. These vector representations are then passed through a single linear layer to project the text features into the 1024-dimensional joint embedding space. We train the model in a contrastive manner to minimize the symmetric loss. The psuedocode for each batch iteration is provided in Algorithm \ref{alg1}.

\raggedbottom
\begin{algorithm}[t!]
  
  \textbf{Input: 1)}A mini-batch of image regions $I_{n}$ with shape ($m \times 3 \times 224 \times 224$) where $m$ is the batch size, and the last 3 dimensions correspond to the image channels, height and width respectively. \textbf{2)}A mini-batch of matching noun pairs $T_{n}$ with shape ($m \times 300$) where 300 represents the dimensions of the GLoVe vectors.
  \textbf{Output:} Symmetric loss for the mini-batch.\\
  \textbf{Initialization:} Pretrained ViT Model with linear head for the image encoder, and a single linear layer for the text encoder.

  \begin{algorithmic}[1]
    \State $h_n = \textnormal{VisionTransformer}(I_n)$ 
    \State  $I_{e} = \textnormal{tanh}(\textbf{W}_{i}\textbf{h}_{n} + \textbf{b}_{i})$ \algorithmiccomment{image embeddings; shape = [m,1024]}
    \State $T_{e} = \textnormal{tanh}(\textbf{W}_{t}T_{n} + \textbf{b}_{t})$ \algorithmiccomment{text embeddings; shape = [m,1024]}
    \State $\textnormal{logits} = T_{e} \times I_{e}^{T}$ \algorithmiccomment{shape = [m, m]}
    \State $I_{sim} = I_{e} \times I_{e}^{T}$ \algorithmiccomment{shape = [m, m]}
    \State $T_{sim} = T_{e} \times T_{e}^T$  \algorithmiccomment{shape = [m, m]}
    \State $\textnormal{labels} = (I_{sim} + T_{sim})/2$ \algorithmiccomment{shape = [m, m]}
    \State $\mathcal{L}_{image} = \textnormal{cross\_entropy\_loss}(\textnormal{labels}^T, \textnormal{logits}^T)$
    \State $\mathcal{L}_{text} = \textnormal{cross\_entropy\_loss}(\textnormal{labels}, \textnormal{logits})$
    \State $\mathcal{L}_{symmetric} = (\mathcal{L}_{image} + \mathcal{L}_{text})/2$
  \end{algorithmic}
    \caption{RoViST-VG}\label{alg1}
\end{algorithm}

\raggedbottom

To compute the visual grounding score, we extract all nouns from the output story sentences and the top 10 bounding box regions for each image in the story based on the confidence scores generated from Faster R-CNN \citep{ren2015faster}. This results in 50 regions for a 5-image story. Each extracted noun and image region is fed through our trained text and image encoder respectively to obtain the image and text embeddings which we denote by $I_{e}$ and $T_{e}$. For each noun, the cosine similarity ($cos$) is calculated between its text embedding with all other region image embeddings. It is noted that a noun mention from a sentence can match with a region from other images and not necessarily just with regions from its corresponding image as we find that words in story sentences may refer to concepts in other images of the sequence. We then use a greedy matching approach to obtain the maximum similarity score for each noun. Following \citet{zhang2019bertscore}, we further experiment by multiplying the similarity score by the inverse document frequency (idf) of the noun calculated from the corpus. This is to put less emphasis on abstract nouns that are not visually grounding but frequently occur in stories (such as `\textit{time}' and `\textit{today}'). Given $N$ stories, the idf score of a token $T_{i}$ is: 

\begin{equation}
    idf(T_{i}) = log(\frac{N}{1+df(T_{i})})
\end{equation}

\noindent where $df(T_{i})$ is the number of stories containing token $T_{i}$. Finally, inspired by \citet{lee2018stacked}, a recall score is computed by using LogSumExp (LSE) pooling:

\begin{equation}
    S_{VG} = \textnormal{log} \sum_{i=1}^{|T_{e}|} \textnormal{exp}(idf(T_{e,i}) \max_{I_{e,j} \in I_{e}}(cos(T_{e,i},I_{e,j})))
\end{equation}

For interpretability, one can optionally scale the score between 0 and 1 using a shifted and scaled version of the sigmoid function: 

\begin{equation}
    S_{VG (scaled)} = \frac{1}{1+\exp{(-0.5 \times S_{VG})}} \times 2 - 1
\end{equation}


\subsection{RoViST-C: Coherence Scorer}
To measure the story's inter-sentence coherence, we leverage the ALBERT model to perform sentence order prediction (SOP) \citep{lan2019albert}. The SOP task is a binary classification task, whereby positive samples are consecutive sentences while negative samples are simply constructed by swapping the two sentences around. This forces the model to primarily focus on learning coherence properties rather than topic prediction. We fine-tune the ALBERT model with adjacent story sentences extracted from the VIST and ROCStories dataset. In total, 822,920 training samples were created where 15\% was used in the validation split. 


Let $\{{s_{i-1}, s_{i}}\}^{N}_{i=1}$ denote the training data where $s_{i-1}$ and $s_{i}$ are adjacent segments. The input sequence fed into ALBERT is in the format $\textbf{s}_{n}$ = `[CLS], $s_{i-1}$, [SEP], $s_{i}$, [SEP]', where [CLS] and [SEP] are special tokens. Then, the pooled $1024$-dimensional vector representation $\textbf{h}_{n}$ of the input sequence is obtained by the output of ALBERT: 

\begin{equation}
    \textbf{h}_{n} = \textnormal{ALBERT} (\textbf{s}_{n})
\end{equation}

To perform SOP, we add a task-specific linear layer on top of ALBERT to predict the probability that $s_{i}$ follows $s_{i-1}$: 

\begin{equation}
    \hat{p}_{n} = \textnormal{softmax}(\textbf{W}_{c}\textbf{h}_{n} + \textbf{b}_{c})
\end{equation}

\noindent where $\textbf{W}_{c}$ and $\textbf{b}_{c}$ are the trainable weights and bias. For the loss function, we optimize the binary cross-entopy loss as follows: 

\begin{equation}
    \mathcal{L} = -p_{n}\textnormal{log}(\hat{p}_{n}) - (1-p_{n})\textnormal{log}(1-\hat{p}_{n})
\end{equation}

To obtain the final coherence score for each story, we compute $\hat{p}_{n}$ for each adjacent sentence pair in the story and average the probabilities across all sentence pairs. 




\subsection{RoViST-NR: Non-redundancy Scorer}
A common problem faced by system output stories is redundancy of words in the form of whole sentences or phrases. While existing methods \citep{hu2020makes} for assessing word diversity and repetition do consider inter-sentence repetition, they do not address repetition \textit{within} sentences. Therefore, to calculate the inter- and intra-sentence non-redundancy score, we propose calculating the Jaccard Similarity (JS) between and within sentences. The JS is defined as the intersection size divided by the union size of two sets \citep{singh2021text}. That is, in our problem, the intersection would be the number of co-occurring words between two texts, while the union is the total number of words in both texts. In particular, we compute the Jaccard Similarity with sentence $\hat{y}_i$ and all its preceding sentences $\{\hat{y}_1,...,\hat{y}_{i-1}\}$ as in Eq. \ref{JS}. Here, $C(\hat{y}_i)$ and $C(\hat{y}_j)$ are the count of unique words in sentence $\hat{y}_i$ and $\hat{y}_j$ respectively. The inter-sentence repetition score is then just simply the average JS scores across the ${n \choose 2}$ sentence pairs where $n$ is the number of sentences in the story.

\begin{equation} \label{JS}
    JS(\hat{y}_i, \hat{y}_j) = \frac{C(\hat{y}_i) \cap C(\hat{y}_j)}{C(\hat{y}_i) \cup C(\hat{y}_j)}
\end{equation}

We also measure the intra-sentence redundancy by first splitting each sentence into non-overlapping \textit{n}-grams and then calculating the JS score between consecutive \textit{n}-grams within sentences. The intra-sentence repetition score for a story is then the average JS scores across all consecutive \textit{n}-gram computations. Lastly, we take the mean of the final inter- and intra-sentence score to obtain the final repetition score for the story and subtract from 1. The result is a score between 0 and 1 where a value closer to 1 means that the story tends to contain less redundancy. 

\section{Data}
\subsection{Supporting Datasets}
\textbf{VIST} The Visual Storytelling Dataset (VIST) dataset \citep{huang2016visual} consists of 10,117 Flickr albums and 210,819 unique images. Each sample is one sequence of 5 photos selected from the same album paired with a single human constructed story, where each story is comprised of mostly one sentence per image. 

\noindent \textbf{ROCStories Corpora} \citep{mostafazadeh2016corpus} is used as additional data along with VIST to train the ALBERT model. It contains 98,161 stories where each story consists of 5 sentences written by humans after being given a prompt. 

\noindent \textbf{Flickr30K Entities} \citep{plummer2015flickr30k} is derived from the Flickr30K dataset \citep{young2014image}, consisting of 31,783 images each matched with 5 captions. The dataset links distinct sentence entities (i.e. a noun/noun phrase) to image bounding boxes, resulting in 70K unique entities and 276K unique bounding boxes. We use the Flickr30K Entities data to train our visual grounding scorer. After filtering out stopwords from the entity mention, we obtained 566K unique entity-region pairs.

\subsection{VST Models}
We evaluate our proposed metric on the output stories produced by 4 state-of-the art VST models: 
1) \textbf{AREL \citep{wang2018no}}: adopts an inverse reinforcement learning approach trained adversarially. The policy model is a CNN+GRU that generates sub-stories for each image, while the reward model is a CNN-based model designed to output the story reward. 2) \textbf{GLACNet \citep{kim2018glac}}: combines both local and global attention. Image features are fed sequentially to a bi-LSTM where the output is a global representation of the entire story. This is concatenated with local image-specific features to create \textit{glocal} vectors which are passed to a decoder for story generation. 3) \textbf{KG-Story \citep{hsu2020knowledge}}: For each image, a word-form conceptual representation is created by predicting a set of terms which are then used to query Visual Genome \citep{krishna2017visual} and OpenIE \citep{pal2016demonyms} to identify links between sets of terms across images. Finally, a Transformer \citep{vaswani2017attention} takes in the term paths to decode the story. 4) \textbf{MCSM+BART \citep{chen2021commonsense}}: image concepts and related concepts extracted from ConceptNet \citep{liu2004conceptnet} are used as input for generating richer stories with BART \citep{lewis2020bart}. To incorporate the most appropriate concepts, their Maximal Clique Selection Module model learns a correlation map, reflecting co-occurrence probabilities of all candidate concepts. 

\section{Evaluation Setup\footnote{The implementation details can be found in the Appendix}}
\begin{table*}[h]
  \centering
  
  \resizebox{0.85\linewidth}{!}{
  \begin{tabular}{@{}cccc|ccc|ccc|ccc@{}}
    \toprule
      & \multicolumn{3}{c|}{\textbf{Grounding}}  &  \multicolumn{3}{c|}{\textbf{Coherence}} & \multicolumn{3}{c|}{\textbf{Non-redun}} & \multicolumn{3}{c}{\textbf{Overall}}  \\
    \midrule
     & $\rho$ & $r$ & $\tau$& $\rho$ & $r$ & $\tau$ & $\rho$ & $r$ & $\tau$ & $\rho$ & $r$ & $\tau$\\ \midrule
    BLEU-1 & 0.198 & 0.168 & 0.127 & 0.052 & 0.044 & 0.030 & 0.018 & -0.044 & 0.019 & 0.080 & 0.064 & 0.051 \\
    BLEU-2 & 0.261 & 0.233 & 0.181 & 0.057 & 0.057 & 0.037 & -0.028 & -0.148 & -0.006 & 0.066 & 0.035 & 0.049 \\
    BLEU-3 & 0.259 & 0.229 & 0.173 & 0.121 & 0.160 & 0.083 & -0.073 & -0.165 & -0.053 & 0.065 & 0.062 & 0.043\\
    BLEU-4 & 0.225 & 0.134 & 0.148 & 0.121 & 0.082 & 0.077 & -0.075 & -0.195 & -0.058 & 0.051 & -0.027 & 0.026 \\
    ROUGE-L & 0.244 & 0.222 & 0.164 & 0.197 & 0.161 & 0.127 & -0.039 & -0.138 & -0.021 & 0.109 & 0.075 & 0.077 \\
    METEOR & 0.348 & 0.319 & 0.228 & 0.291 & 0.256 & 0.213 & 0.203 & 0.075 & 0.140 & 0.327 & 0.280 & 0.223 \\
    CIDEr & 0.269 & 0.158 & 0.194 & 0.207 & 0.104 & 0.146 & 0.013 & -0.190 & 0.005 & 0.182 & -0.005  & 0.131 \\
    SPICE & 0.311 & 0.301 & 0.214 & 0.052 & 0.069 & 0.031 & 0.018 & -0.051 & 0.015 & 0.127 & 0.134 & 0.095 \\
    WMD & 0.472 & 0.490 & 0.337 & 0.186 & 0.236 & 0.129 & 0.106 & 0.015 & 0.076 & 0.262 & 0.312 & 0.183 \\
    $F_{BERT}$ & 0.180 & 0.175  & 0.149 & 0.287 & 0.320 & 0.202 & 0.088 & 0.038 & 0.061 & 0.199 & 0.218 &  0.128 \\
    TIGEr & \textbf{0.519} & \textbf{0.504} & 0.354 & -0.03 & -0.089 & -0.027 & -0.224 & -0.325 & -0.147 & 0.010 & -0.005 & 0.010 \\ \midrule
    RoViST(-VG/C/NR) & 0.509 & 0.460 & \textbf{0.365} & \textbf{0.446} & \textbf{0.456} & \textbf{0.308} & \textbf{0.531} & \textbf{0.736}& \textbf{0.397} & \textbf{0.554}  & \textbf{0.579} & \textbf{0.387}  \\
    \bottomrule
  \end{tabular}}
  \caption{Criteria level Spearman’s $\rho$, Pearson’s $r$ and Kendall's $\tau$ correlations between automatic metrics and mean of human scores. Correlations for Grounding, Coherence, Non-redun and Overall are measured with RoViST-VG, RoViST-C, RoViST-NR and RoViST respectively.} \label{criteria correls} 
\end{table*}

\textbf{Evaluation Metrics}
To assess the performance for RoViST, we analyze its correlation with reliable human judgements by recruiting many responders (26) whereas related works \citep{guan2020union, hu2020makes} have used 3-7 annotators. In total, the 26 responders analysed 400 machine generated sentences across 80 stories and 4 models, including AREL, GLACNet, KG-Story and MCSM+BART. A Likert scale was used to score 3 different criteria for each story based on what we believe defines a good story - 1) the story is visually grounded, 2) sentences are natural sounding and topically coherent, and 3) there is no repeating plots within the story. Annotators were additionally asked to vote for which of the 4 models produced the best story relating to the visual prompt based on no particular criteria. We follow existing literature and report the Spearman's correlation $\rho$, Pearson's correlation $r$ and Kendall's correlation $\tau$.
\\
\\
\noindent \textbf{Baseline} 
We select 11 baseline metrics to compare with our metric: BLEU-1,2,3,4 \citep{papineni2002bleu}, ROUGE-L \citep{lin2004rouge}, METEOR \citep{banerjee2005meteor}, CIDEr \citep{vedantam2015cider}, SPICE \citep{anderson2016spice}, WMD \citep{kusner2015word}, $F_{BERT}$ (F1-measure version of BERTScore) \citep{zhang2019bertscore} and TIGEr \citep{jiang2019tiger}.

\section{Results}

\subsection{Human Scores versus Story Ranking}

 \begin{figure}[t!]
  \centering
  \includegraphics[width=1\linewidth]{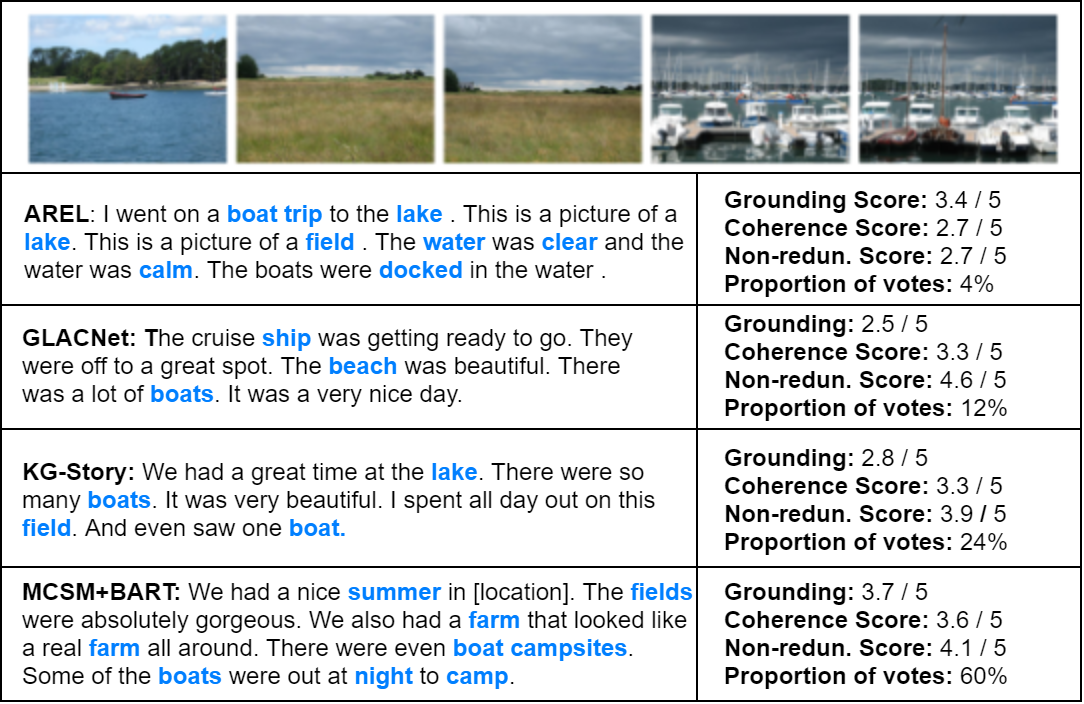}
  \caption{Average human scores for an example story across 3 criteria for 4 different VST models. `Proportion of votes' refers to the percentage of voters who voted that model's story as the best out of the 4. Blue highlighted words visually relate to the image.}
  \label{fig: human score}
\end{figure}

 We first investigate whether there is any correlation between the human scores for each 3 criteria and the model that was voted as the best for each photo sequence. For each photo sequence, we rank each of the 4 models' stories based on the proportion of votes that it received. The correlations were then calculated between the mean human scores for each criteria and the model rankings, and the average correlation coefficients were finally taken across the unique stories to obtain the values in Table \ref{human criteria correl}. We also sum up the human scores across the 3 criteria and measure its correlation with the rankings to further analyze at an \textit{Overall} level.
 
  \begin{table}[t!]
  \centering
  \resizebox{.6\linewidth}{!}{
  \begin{tabular}{@{}cccc}
    \toprule
     & $\rho$ & $r$ & $\tau$ \\
    \midrule
     Grounding & 0.423 & 0.434 & 0.400 \\ 
     Coherence & 0.663 & 0.698 &  0.618 \\ 
     Non-redun & 0.379 & 0.484 & 0.328 \\ \midrule
     Overall & \textbf{0.754} & \textbf{0.769} & \textbf{0.676} \\ \bottomrule
  \end{tabular}}
  \caption{Criteria level Spearman’s $\rho$, Pearson’s $r$ and Kendall's $\tau$ between human scores and story ranking.} \label{human criteria correl} 
\end{table}

 Interestingly, we find that sentence coherence plays the most significant role when ranking stories whereas non-redundancy and visual grounding are less important. Figure \ref{fig: human score} provides an example of this case where our human annotators preferred KG-Story and GLACNet over AREL which was more visually grounding but less coherent-sounding. We observe even stronger correlation when we sum the 3 criteria scores, suggesting that all 3 aspects combined can give better guidance when judging a story as can be seen in Figure \ref{fig: human score} where most of the votes went to MCSM+BART which scored relatively well in all 3 areas.

\subsection{Correlation Analysis with Human Scores} 
Table \ref{criteria correls} displays the correlation between the metrics and the mean human scores. The results were analyzed at a criteria level by examining correlations between each criteria's scores with our  metric which targets that criteria. We also analyze the \textit{Overall} scores by summing up the 3 criteria scores and measuring its correlation with RoViST which represents the sum of the scores produced by RoViST-VG, RoViST-C and RoViST-NR. 

With the grounding correlations, RoViST-VG outperforms the baselines for Kendall's correlation. However, it is slightly outperformed by TIGEr when comparing Spearman's correlation and by TIGEr and WMD when comparing Pearson's correlation. We note that all baseline metrics are reference-based and therefore, a likely explanation for the moderate correlations for even simple metrics like METEOR is that human references can already provide a good guideline when assessing text-to-image relatedness. Moreover, we hypothesize that image captioning metrics will perform well for the visual grounding aspect in the case when the model happens to output a sentence that sounds like an image caption. However, unlike image captioning, we emphasize that just having high correlation between image objects and text descriptions does not necessarily mean a good story as we highlighted in the previous section. Examining the coherence and non-redundancy aspect, we observe that a majority of the baselines correlate poorly. Conversely, our RoViST-C and RoViST-NR metric designed to specifically target these criteria generated significantly higher correlations. When comparing at the \textit{Overall} level, we also achieved noticeably better results in terms of $\rho$, $r$ and $\tau$. 


 
\subsection{Changing Number of References}

\begin{figure}[t!]
  \centering
  \includegraphics[width=1\linewidth]{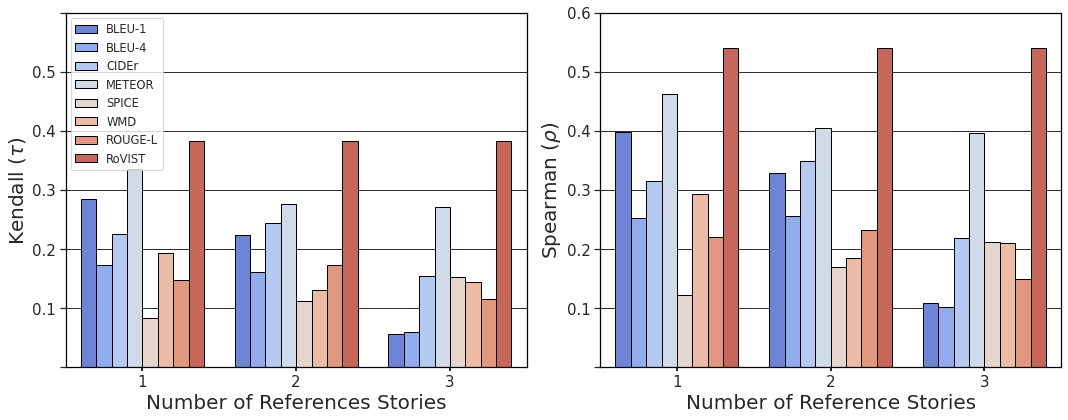}
  \caption{Kendall (left) and Spearman (right) correlation vs. Number of References.}
  \label{fig: spearman kendall ref}
\end{figure}

Figure \ref{fig: spearman kendall ref} shows how the Spearman and Kendall correlations for some of the metrics vary with different number of human-written references versus our reference-less metric. The stories selected for our analysis each have a different number of reference stories ranging from 1 to 4. As there were not many stories with 4 references, we select those stories that had 3 references, resulting in 60 stories with 300 sentences for analysis. We then compute the correlations with the human judgement across the metrics using 1,2, and 3 references. 

It is evident that the results from the reference-based metrics fluctuate significantly according to the number of references. However, the trend is unclear. Increasing the number of references from 1 to 2 appears to improve the correlations for some of the metrics like CIDEr, SPICE and ROUGE-L. This may be because having more references can better capture allowable variations in storytelling compared to a single reference. However, increasing from 2 to 3 references actually worsens the performance for many of the metrics (BLEU-1/4, CIDEr, METEOR and ROUGE-L). A possible explanation could be that the additional reference added may have caused bias for some metrics. In particular, $n$-gram based metrics like BLEU and ROUGE focus on $n$-gram overlap. Thus, it is possible that the additional reference introduced may have a high $n$-gram overlap with the candidate but for unimportant filler words like `\textit{the}' or `\textit{and}'. Our metric on the other hand, alleviates this issue by first being a reference-free metric and secondly, by only focusing on important words (nouns) in the candidate story via RoViST-VG.

It is also noted that examining more amount of references could potentially reveal a better trend. However, this is challenging as the maximum amount of references in the VIST dataset is 5 with 82.50\% of the stories having 3 or less. Moreover, collecting multiple human reference stories is an expensive process in most cases.  

\begin{figure}[t!]
  \centering
  \includegraphics[width=1\linewidth]{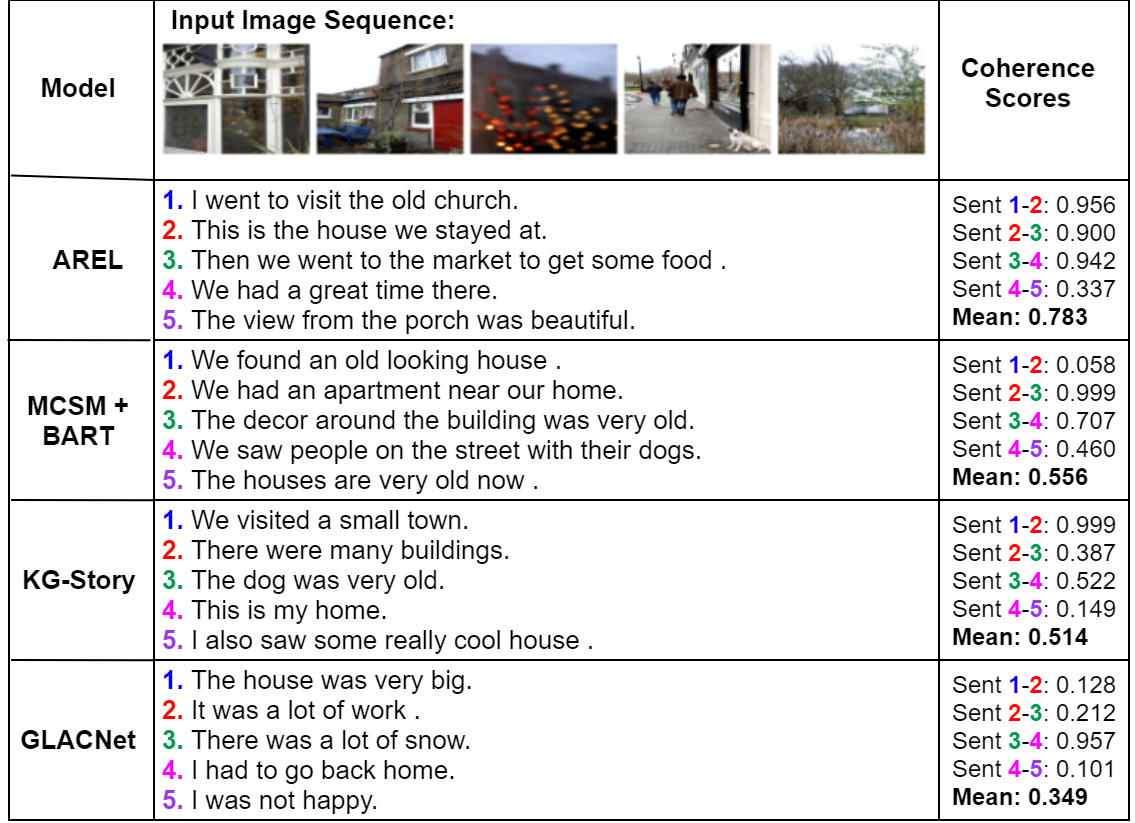}
  \caption{Predicted coherence probabilities from RoViST-C for 4 VST models.}
  \label{fig: qualitative albert}
\end{figure}

\subsection{Qualitative Analysis}
We conduct qualitative analysis on our visual grounding scorer (RoViST-VG) and coherence scorer (RoViST-C).

 \textbf{RoViST-VG} Figure \ref{fig: qualitative clip} in Appendix A displays an example gold story with noun mentions highlighted in blue, followed by the corresponding bounding box regions that gave the highest similarity score retrieved by our RoViST-VG model. We observe that the model performs well at matching a majority of the nouns. However, words that are less visually grounding like `\textit{corner}' or intangible nouns such as `\textit{visit}' are extremely challenging to ground. Consequently, RoViST-VG can sometimes retrieve a region that is not closely related for these types of words. This also occurs for words that are mentioned in the story but not explicitly shown in the images like the word `\textit{photos}' in Example Story 2. A potential problem of this may be the presence of false positives if a story tends to mention many non-visual entities. This could lead to a higher grounding score compared to a story that only mentions a few entities that are visually grounded. Nevertheless, our model still serves as guidance for analyzing how visually detailed a story is and can also reflect how many related entities are mentioned.

\textbf{RoViST-C} The qualititive results for 4 example machine stories is displayed in Figure \ref{fig: qualitative albert}. Noticeably, RoViST-C tends to assign higher probabilities to sentences that flow. These sentences do not necessarily need to be about the same topic. For instance, sentence 2 and 3 in AREL's story each have a different topic focus but the sentence transition is given a 0.90 coherence score as they follow a narrative style. Conversely, consecutive sentences with similar topics but are incoherent can be given low scores such as sentences 4-5 from KG-Story. It is clear that training ALBERT with sentence order prediction allows the model to capture inter-sentence coherence and is not just limited to modelling topic similarity across sentences. 

\section{Conclusion}
We propose RoViST, a metric for evaluating VST tasks on 3 aspects: visual grounding, coherence and non-redundancy. RoViST correlates well with human judgement, outperforming other metrics when comparing the coherence and non-redundancy criteria as well as when combining all 3 criteria. While some existing metrics slightly outperform our method on visual grounding, we note that image-to-text similarity is just one aspect of VST and this aspect alone is insufficient in defining a good story. Unlike other metrics, RoViST is reference-free and hence, robust to the number of references which are costly to obtain for VST. It is also interpretable and can be used to gauge out where the model is underperforming. We hope that RoViST provides preliminary insight into future work on developing VST models and evaluations. 


\bibliography{custom}

\begin{thebibliography}{40}
\expandafter\ifx\csname natexlab\endcsname\relax\def\natexlab#1{#1}\fi

\bibitem[{Anderson et~al.(2016)Anderson, Fernando, Johnson, and
  Gould}]{anderson2016spice}
Peter Anderson, Basura Fernando, Mark Johnson, and Stephen Gould. 2016.
\newblock Spice: Semantic propositional image caption evaluation.
\newblock In \emph{European conference on computer vision}, pages 382--398.
  Springer.

\bibitem[{Anderson et~al.(2018)Anderson, He, Buehler, Teney, Johnson, Gould,
  and Zhang}]{anderson2018bottom}
Peter Anderson, Xiaodong He, Chris Buehler, Damien Teney, Mark Johnson, Stephen
  Gould, and Lei Zhang. 2018.
\newblock Bottom-up and top-down attention for image captioning and visual
  question answering.
\newblock In \emph{Proceedings of the IEEE conference on computer vision and
  pattern recognition}, pages 6077--6086.

\bibitem[{Banerjee and Lavie(2005)}]{banerjee2005meteor}
Satanjeev Banerjee and Alon Lavie. 2005.
\newblock Meteor: An automatic metric for mt evaluation with improved
  correlation with human judgments.
\newblock In \emph{Proceedings of the acl workshop on intrinsic and extrinsic
  evaluation measures for machine translation and/or summarization}, pages
  65--72.

\bibitem[{Chen et~al.(2021)Chen, Huang, Takamura, and
  Nakayama}]{chen2021commonsense}
Hong Chen, Yifei Huang, Hiroya Takamura, and Hideki Nakayama. 2021.
\newblock Commonsense knowledge aware concept selection for diverse and
  informative visual storytelling.
\newblock In \emph{Proceedings of the AAAI Conference on Artificial
  Intelligence}, volume~35, pages 999--1008.

\bibitem[{Cui et~al.(2018)Cui, Yang, Veit, Huang, and
  Belongie}]{cui2018learning}
Yin Cui, Guandao Yang, Andreas Veit, Xun Huang, and Serge Belongie. 2018.
\newblock Learning to evaluate image captioning.
\newblock In \emph{Proceedings of the IEEE conference on computer vision and
  pattern recognition}, pages 5804--5812.

\bibitem[{Devlin et~al.(2018)Devlin, Chang, Lee, and
  Toutanova}]{devlin2018bert}
Jacob Devlin, Ming-Wei Chang, Kenton Lee, and Kristina Toutanova. 2018.
\newblock Bert: Pre-training of deep bidirectional transformers for language
  understanding.

\bibitem[{Dosovitskiy et~al.(2020)Dosovitskiy, Beyer, Kolesnikov, Weissenborn,
  Zhai, Unterthiner, Dehghani, Minderer, Heigold, Gelly
  et~al.}]{dosovitskiy2020image}
Alexey Dosovitskiy, Lucas Beyer, Alexander Kolesnikov, Dirk Weissenborn,
  Xiaohua Zhai, Thomas Unterthiner, Mostafa Dehghani, Matthias Minderer, Georg
  Heigold, Sylvain Gelly, et~al. 2020.
\newblock An image is worth 16$\times$ 16 words: Transformers for image
  recognition at scale.

\bibitem[{Guan and Huang(2020)}]{guan2020union}
Jian Guan and Minlie Huang. 2020.
\newblock Union: An unreferenced metric for evaluating open-ended story
  generation.
\newblock In \emph{Proceedings of the 2020 Conference on Empirical Methods in
  Natural Language Processing (EMNLP)}, pages 9157--9166.

\bibitem[{Hsu et~al.(2020)Hsu, Chen, Hsu, Li, Lin, Huang, and
  Ku}]{hsu2020knowledge}
Chao-Chun Hsu, Zi-Yuan Chen, Chi-Yang Hsu, Chih-Chia Li, Tzu-Yuan Lin, Ting-Hao
  Huang, and Lun-Wei Ku. 2020.
\newblock Knowledge-enriched visual storytelling.
\newblock In \emph{Proceedings of the AAAI Conference on Artificial
  Intelligence}, volume~34, pages 7952--7960.

\bibitem[{Hsu et~al.(2019)Hsu, Huang, Hsu, and Huang}]{hsu2019visual}
Ting-Yao Hsu, Chieh-Yang Huang, Yen-Chia Hsu, and Ting-Hao Huang. 2019.
\newblock Visual story post-editing.
\newblock In \emph{Proceedings of the 57th Annual Meeting of the Association
  for Computational Linguistics}, pages 6581--6586.

\bibitem[{Hu et~al.(2020)Hu, Cheng, Gan, Liu, Gao, and Neubig}]{hu2020makes}
Junjie Hu, Yu~Cheng, Zhe Gan, Jingjing Liu, Jianfeng Gao, and Graham Neubig.
  2020.
\newblock What makes a good story? designing composite rewards for visual
  storytelling.
\newblock In \emph{Proceedings of the AAAI Conference on Artificial
  Intelligence}, volume~34, pages 7969--7976.

\bibitem[{Huang et~al.(2016)Huang, Ferraro, Mostafazadeh, Misra, Agrawal,
  Devlin, Girshick, He, Kohli, Batra et~al.}]{huang2016visual}
Ting-Hao Huang, Francis Ferraro, Nasrin Mostafazadeh, Ishan Misra, Aishwarya
  Agrawal, Jacob Devlin, Ross Girshick, Xiaodong He, Pushmeet Kohli, Dhruv
  Batra, et~al. 2016.
\newblock Visual storytelling.
\newblock In \emph{Proceedings of the 2016 Conference of the North American
  Chapter of the Association for Computational Linguistics: Human Language
  Technologies}, pages 1233--1239.

\bibitem[{Jiang et~al.(2019)Jiang, Huang, Zhang, Wang, Zhang, Gan, Diesner, and
  Gao}]{jiang2019tiger}
Ming Jiang, Qiuyuan Huang, Lei Zhang, Xin Wang, Pengchuan Zhang, Zhe Gan, Jana
  Diesner, and Jianfeng Gao. 2019.
\newblock Tiger: Text-to-image grounding for image caption evaluation.
\newblock In \emph{Proceedings of the 2019 Conference on Empirical Methods in
  Natural Language Processing and the 9th International Joint Conference on
  Natural Language Processing (EMNLP-IJCNLP)}, pages 2141--2152.

\bibitem[{Kim et~al.(2018)Kim, Heo, Son, Park, and Zhang}]{kim2018glac}
Taehyeong Kim, Min-Oh Heo, Seonil Son, Kyoung-Wha Park, and Byoung-Tak Zhang.
  2018.
\newblock Glac net: Glocal attention cascading networks for multi-image cued
  story generation.

\bibitem[{Kingma and Ba(2014)}]{kingma2014adam}
Diederik~P Kingma and Jimmy Ba. 2014.
\newblock Adam: A method for stochastic optimization.

\bibitem[{Krishna et~al.(2017)Krishna, Zhu, Groth, Johnson, Hata, Kravitz,
  Chen, Kalantidis, Li, Shamma et~al.}]{krishna2017visual}
Ranjay Krishna, Yuke Zhu, Oliver Groth, Justin Johnson, Kenji Hata, Joshua
  Kravitz, Stephanie Chen, Yannis Kalantidis, Li-Jia Li, David~A Shamma, et~al.
  2017.
\newblock Visual genome: Connecting language and vision using crowdsourced
  dense image annotations.
\newblock \emph{International journal of computer vision}, 123(1):32--73.

\bibitem[{Kusner et~al.(2015)Kusner, Sun, Kolkin, and
  Weinberger}]{kusner2015word}
Matt Kusner, Yu~Sun, Nicholas Kolkin, and Kilian Weinberger. 2015.
\newblock From word embeddings to document distances.
\newblock In \emph{International conference on machine learning}, pages
  957--966. PMLR.

\bibitem[{Lan et~al.(2019)Lan, Chen, Goodman, Gimpel, Sharma, and
  Soricut}]{lan2019albert}
Zhenzhong Lan, Mingda Chen, Sebastian Goodman, Kevin Gimpel, Piyush Sharma, and
  Radu Soricut. 2019.
\newblock Albert: A lite bert for self-supervised learning of language
  representations.
\newblock In \emph{International Conference on Learning Representations}.

\bibitem[{Lee et~al.(2020)Lee, Yoon, Dernoncourt, Kim, Bui, and
  Jung}]{lee2020vilbertscore}
Hwanhee Lee, Seunghyun Yoon, Franck Dernoncourt, Doo~Soon Kim, Trung Bui, and
  Kyomin Jung. 2020.
\newblock Vilbertscore: Evaluating image caption using vision-and-language
  bert.
\newblock In \emph{Proceedings of the First Workshop on Evaluation and
  Comparison of NLP Systems}, pages 34--39.

\bibitem[{Lee et~al.(2018)Lee, Chen, Hua, Hu, and He}]{lee2018stacked}
Kuang-Huei Lee, Xi~Chen, Gang Hua, Houdong Hu, and Xiaodong He. 2018.
\newblock Stacked cross attention for image-text matching.
\newblock In \emph{Proceedings of the European Conference on Computer Vision
  (ECCV)}, pages 201--216.

\bibitem[{Lewis et~al.(2020)Lewis, Liu, Goyal, Ghazvininejad, Mohamed, Levy,
  Stoyanov, and Zettlemoyer}]{lewis2020bart}
Mike Lewis, Yinhan Liu, Naman Goyal, Marjan Ghazvininejad, Abdelrahman Mohamed,
  Omer Levy, Veselin Stoyanov, and Luke Zettlemoyer. 2020.
\newblock Bart: Denoising sequence-to-sequence pre-training for natural
  language generation, translation, and comprehension.
\newblock In \emph{Proceedings of the 58th Annual Meeting of the Association
  for Computational Linguistics}, pages 7871--7880.

\bibitem[{Lin(2004)}]{lin2004rouge}
Chin-Yew Lin. 2004.
\newblock Rouge: A package for automatic evaluation of summaries.
\newblock In \emph{Text summarization branches out}, pages 74--81.

\bibitem[{Liu and Singh(2004)}]{liu2004conceptnet}
Hugo Liu and Push Singh. 2004.
\newblock Conceptnet—a practical commonsense reasoning tool-kit.
\newblock \emph{BT technology journal}, 22(4):211--226.

\bibitem[{Lu et~al.(2019)Lu, Batra, Parikh, and Lee}]{lu2019vilbert}
Jiasen Lu, Dhruv Batra, Devi Parikh, and Stefan Lee. 2019.
\newblock Vilbert: pretraining task-agnostic visiolinguistic representations
  for vision-and-language tasks.
\newblock In \emph{Proceedings of the 33rd International Conference on Neural
  Information Processing Systems}, pages 13--23.

\bibitem[{Mostafazadeh et~al.(2016)Mostafazadeh, Chambers, He, Parikh, Batra,
  Vanderwende, Kohli, and Allen}]{mostafazadeh2016corpus}
Nasrin Mostafazadeh, Nathanael Chambers, Xiaodong He, Devi Parikh, Dhruv Batra,
  Lucy Vanderwende, Pushmeet Kohli, and James Allen. 2016.
\newblock A corpus and cloze evaluation for deeper understanding of commonsense
  stories.
\newblock In \emph{Proceedings of the 2016 Conference of the North American
  Chapter of the Association for Computational Linguistics: Human Language
  Technologies}, pages 839--849.

\bibitem[{Pal et~al.(2016)}]{pal2016demonyms}
Harinder Pal et~al. 2016.
\newblock Demonyms and compound relational nouns in nominal open ie.
\newblock In \emph{Proceedings of the 5th Workshop on Automated Knowledge Base
  Construction}, pages 35--39.

\bibitem[{Papineni et~al.(2002)Papineni, Roukos, Ward, and
  Zhu}]{papineni2002bleu}
Kishore Papineni, Salim Roukos, Todd Ward, and Wei-Jing Zhu. 2002.
\newblock Bleu: a method for automatic evaluation of machine translation.
\newblock In \emph{Proceedings of the 40th annual meeting of the Association
  for Computational Linguistics}, pages 311--318.

\bibitem[{Pennington et~al.(2014)Pennington, Socher, and
  Manning}]{pennington2014glove}
Jeffrey Pennington, Richard Socher, and Christopher~D Manning. 2014.
\newblock Glove: Global vectors for word representation.
\newblock In \emph{Proceedings of the 2014 conference on empirical methods in
  natural language processing (EMNLP)}, pages 1532--1543.

\bibitem[{Plummer et~al.(2015)Plummer, Wang, Cervantes, Caicedo, Hockenmaier,
  and Lazebnik}]{plummer2015flickr30k}
Bryan~A Plummer, Liwei Wang, Chris~M Cervantes, Juan~C Caicedo, Julia
  Hockenmaier, and Svetlana Lazebnik. 2015.
\newblock Flickr30k entities: Collecting region-to-phrase correspondences for
  richer image-to-sentence models.
\newblock In \emph{Proceedings of the IEEE international conference on computer
  vision}, pages 2641--2649.

\bibitem[{Radford et~al.(2021)Radford, Kim, Hallacy, Ramesh, Goh, Agarwal,
  Sastry, Askell, Mishkin, Clark et~al.}]{radford2021learning}
Alec Radford, Jong~Wook Kim, Chris Hallacy, Aditya Ramesh, Gabriel Goh,
  Sandhini Agarwal, Girish Sastry, Amanda Askell, Pamela Mishkin, Jack Clark,
  et~al. 2021.
\newblock Learning transferable visual models from natural language
  supervision.

\bibitem[{Ren et~al.(2015)Ren, He, Girshick, and Sun}]{ren2015faster}
Shaoqing Ren, Kaiming He, Ross Girshick, and Jian Sun. 2015.
\newblock Faster r-cnn: Towards real-time object detection with region proposal
  networks.
\newblock \emph{Advances in neural information processing systems}, 28:91--99.

\bibitem[{Sharif et~al.(2018)Sharif, White, Bennamoun, and
  Shah}]{sharif2018nneval}
Naeha Sharif, Lyndon White, Mohammed Bennamoun, and Syed Afaq~Ali Shah. 2018.
\newblock Nneval: Neural network based evaluation metric for image captioning.
\newblock In \emph{Proceedings of the European Conference on Computer Vision
  (ECCV)}, pages 37--53.

\bibitem[{Singh and Singh(2021)}]{singh2021text}
Ritika Singh and Satwinder Singh. 2021.
\newblock Text similarity measures in news articles by vector space model using
  nlp.
\newblock \emph{Journal of The Institution of Engineers (India): Series B},
  102(2):329--338.

\bibitem[{Vaswani et~al.(2017)Vaswani, Shazeer, Parmar, Uszkoreit, Jones,
  Gomez, Kaiser, and Polosukhin}]{vaswani2017attention}
Ashish Vaswani, Noam Shazeer, Niki Parmar, Jakob Uszkoreit, Llion Jones,
  Aidan~N Gomez, {\L}ukasz Kaiser, and Illia Polosukhin. 2017.
\newblock Attention is all you need.
\newblock In \emph{Advances in neural information processing systems}, pages
  5998--6008.

\bibitem[{Vedantam et~al.(2015)Vedantam, Lawrence~Zitnick, and
  Parikh}]{vedantam2015cider}
Ramakrishna Vedantam, C~Lawrence~Zitnick, and Devi Parikh. 2015.
\newblock Cider: Consensus-based image description evaluation.
\newblock In \emph{Proceedings of the IEEE conference on computer vision and
  pattern recognition}, pages 4566--4575.

\bibitem[{Wang et~al.(2018)Wang, Chen, Wang, and Wang}]{wang2018no}
Xin Wang, Wenhu Chen, Yuan-Fang Wang, and William~Yang Wang. 2018.
\newblock No metrics are perfect: Adversarial reward learning for visual
  storytelling.
\newblock In \emph{Proceedings of the 56th Annual Meeting of the Association
  for Computational Linguistics (Volume 1: Long Papers)}, pages 899--909.

\bibitem[{Young et~al.(2014)Young, Lai, Hodosh, and
  Hockenmaier}]{young2014image}
Peter Young, Alice Lai, Micah Hodosh, and Julia Hockenmaier. 2014.
\newblock From image descriptions to visual denotations: New similarity metrics
  for semantic inference over event descriptions.
\newblock \emph{Transactions of the Association for Computational Linguistics},
  2:67--78.

\bibitem[{Zhang et~al.(2019)Zhang, Kishore, Wu, Weinberger, and
  Artzi}]{zhang2019bertscore}
Tianyi Zhang, Varsha Kishore, Felix Wu, Kilian~Q Weinberger, and Yoav Artzi.
  2019.
\newblock Bertscore: Evaluating text generation with bert.
\newblock In \emph{International Conference on Learning Representations}.

\bibitem[{Zhao et~al.(2019)Zhao, Peyrard, Liu, Gao, Meyer, and
  Eger}]{zhao2019moverscore}
Wei Zhao, Maxime Peyrard, Fei Liu, Yang Gao, Christian~M Meyer, and Steffen
  Eger. 2019.
\newblock Moverscore: Text generation evaluating with contextualized embeddings
  and earth mover distance.
\newblock In \emph{Proceedings of the 2019 Conference on Empirical Methods in
  Natural Language Processing (EMNLP)}.

\bibitem[{Zhou et~al.(2020)Zhou, Palangi, Zhang, Hu, Corso, and
  Gao}]{zhou2020unified}
Luowei Zhou, Hamid Palangi, Lei Zhang, Houdong Hu, Jason Corso, and Jianfeng
  Gao. 2020.
\newblock Unified vision-language pre-training for image captioning and vqa.
\newblock In \emph{Proceedings of the AAAI Conference on Artificial
  Intelligence}, volume~34, pages 13041--13049.

\end{thebibliography}
\bibliographystyle{acl_natbib}

\newpage
\appendix

\section{RoViST-VG Example Output}
\begin{figure*}[b]
  \centering
  \includegraphics[width=0.8\linewidth]{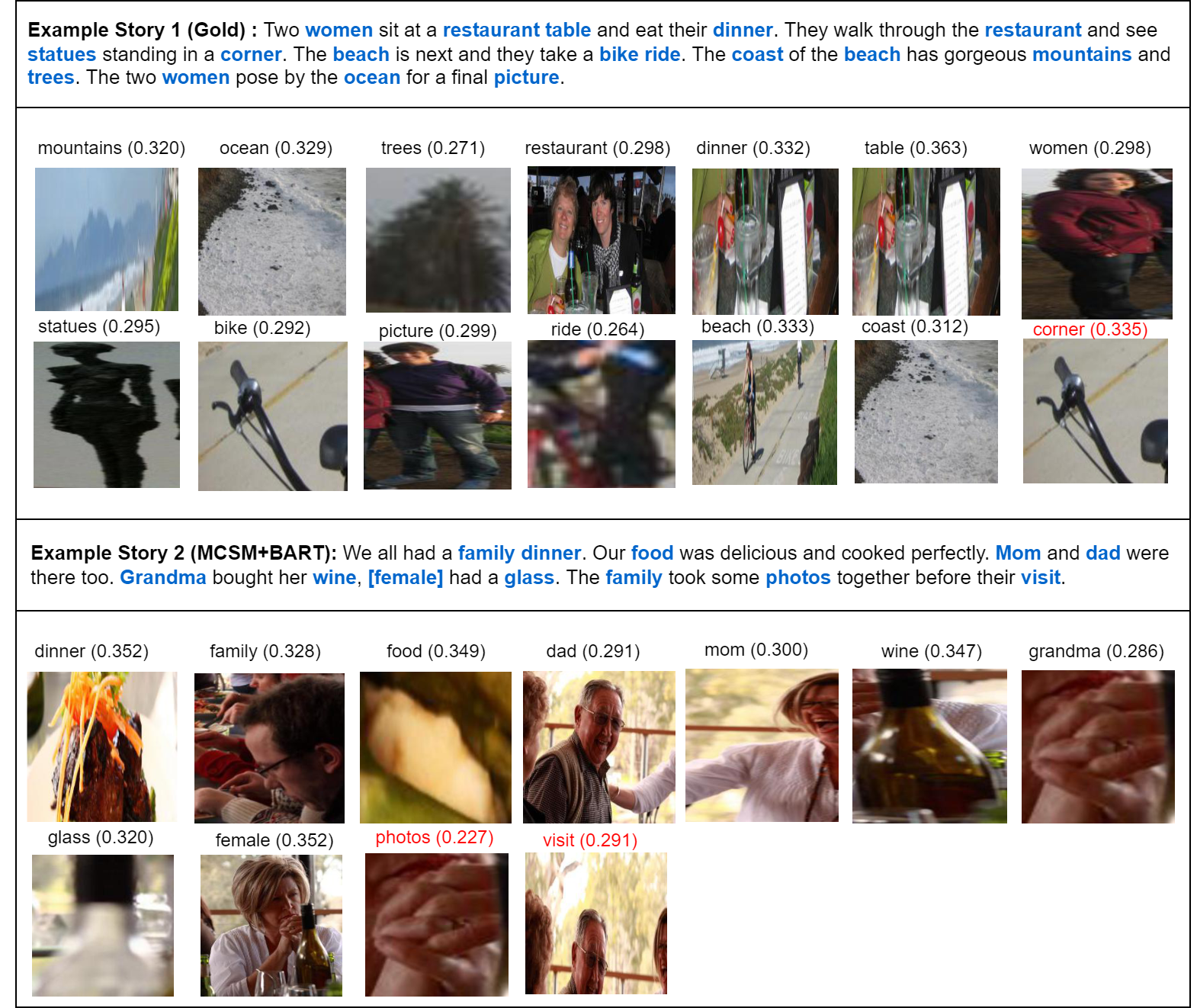}
  \caption{Retrieved regions from RoViST-VG for an example gold story (top) and machine generated story (bottom).}
  \label{fig: qualitative clip}
\end{figure*}

Figure \ref{fig: qualitative clip} shows the retrieved regions from the RoViST-VG model for an example gold story (top) and a machine-story generated from the MCSM+BART model (bottom). The blue highlighted words are the nouns while red highlighted words indicate words that do not explicitly appear in the image sequence or are less visually grounding words.

\section{Implementation Details}

\textbf{RoViST-VG} We use the Adam optimizer \citep{kingma2014adam} with a 0.00001 weight decay. The learning rate was initially set to 0.00005 and was reduced by 5\% with each consecutive epoch. For the ViT model, we use the `\textit{vit-base-patch16-224}' style configuration which outputs image features as a 768 dimensional vector. Further, the linear layer used to project the text and embedding features to the joint embedding space (of dimension 1024) uses a $tanh$ activation function. No normalization of the image and text embeddings was done during the training process as we did not find any benefit from doing this. Finally, we set the mini-batch size to 64 and use early stopping to cease training after the validation loss fails to improve for 3 consecutive epochs. We note that 85\% and 15\% of the data was used in the training and validation set respectively. The model converged in 3 epochs, taking approximately 12 hours with a Nvidia Tesla P100 GPU.
\\
\\
\noindent \textbf{RoViST-C} For ALBERT, we use the `\textit{albert-large-v1}' configuration and the Adam optimizer with a 0.00001 weight decay for training. The learning rate was 0.00001 which we schedule to reduce by 5\% every epoch. Additionally, the batch size was 32 and early stopping was employed after the validation loss failed to improve for 5 epochs. We note that 85\% and 15\% of the data was used in the training and validation set respectively. In total, we trained the model for 5 epochs, taking 14 hours with a Nvidia Tesla P100 GPU.
\\
\\
\noindent \textbf{RoViST-NR} For assessing intra-sentence non-redundancy, $n$-grams of size 4 were used as we found that repetition of words within sentences usually occurred in fours.
\raggedbottom

\onecolumn 

\section{Human Evaluation Survey}
\begin{figure}[H]
  \centering
  \includegraphics[width=1\linewidth]{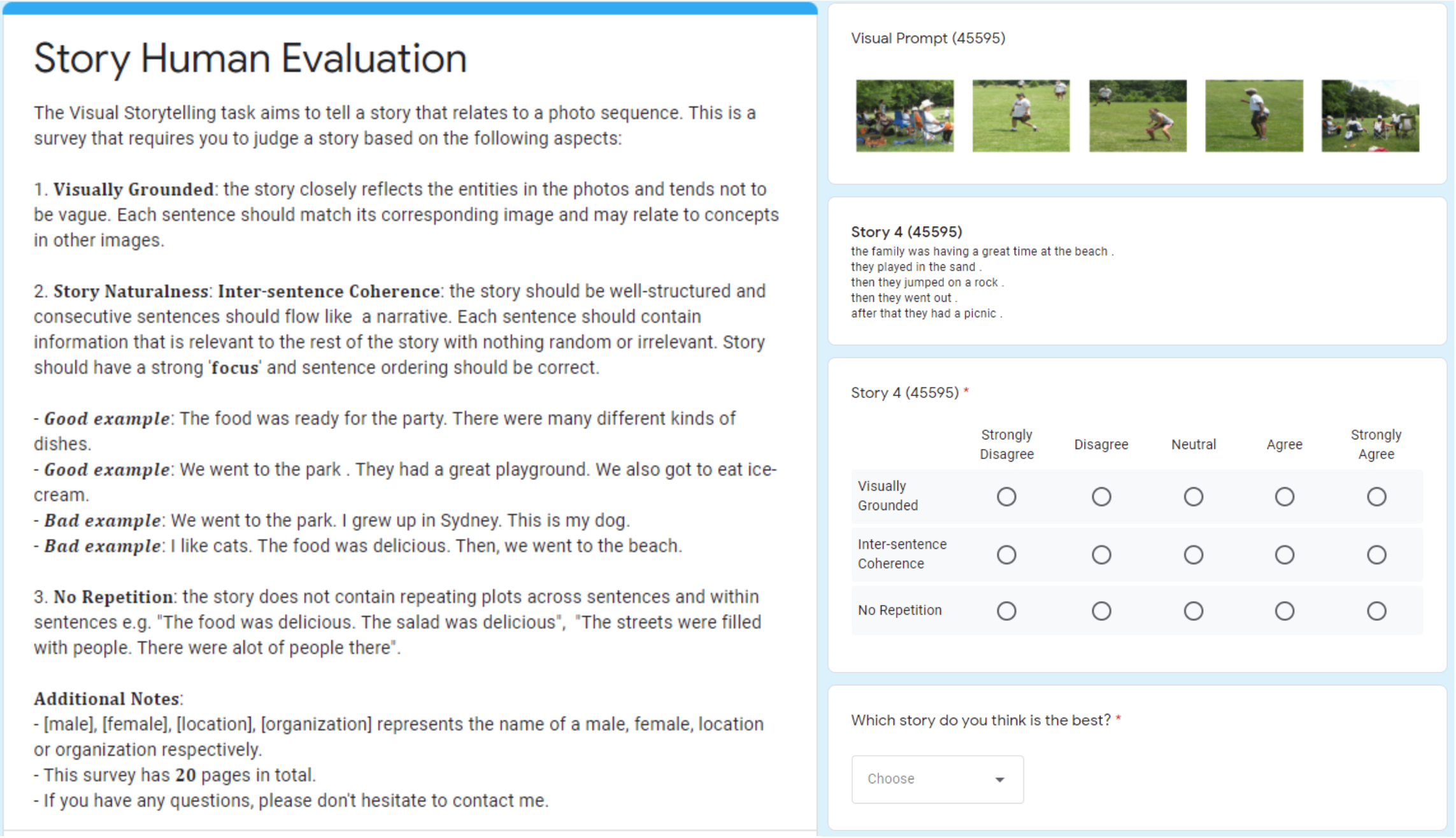}
  \caption{Survey instructions and form format for the human evaluation study. }
  \label{fig: survey}
\end{figure}

Figure \ref{fig: survey} shows the survey instructions used in the human evaluation study and the format of the survey questions. Participants recruited were volunteers from a variety of age groups (20-60 years old), education level and gender (10 female, 16 male).

\end{document}